\documentstyle[epic,eepic,ecltree]{llncs}

\newcommand{\Xcal}[1]{\mbox{${\cal #1}$}}
\newcommand{\Xsf}[1]{\mbox{\sf #1}}
\newcommand{\denI}[1]{\mbox{$ [ \! [  {#1} ] \! ] ^ {\cal I} $}}

\newcommand{\denIV}[1]{\mbox{$ [ \! [  {#1} ] \! ] ^ {\cal I}_V $}}
\newcommand{\denA}[1]{\mbox{$ [ \! [  {#1} ] \! ] ^ {\cal A} $}}
\newcommand{\denAi}[2]{\mbox{$ [ \! [  {#1}  ] \! ] ^ {{\cal A}_{#2}} $}}
\newcommand{\denAV}[1]{\mbox{$ [ \! [  {#1} ] \! ] ^ {\cal A}_V $}}
\newcommand{\denX}[2]{\mbox{$ [ \! [  {#1} ] \! ] ^ {\cal {#2}} $}}
\newcommand{\Pf}{\mbox{${\cal P}_{\rm F}\:$}}
\newcommand{\Rl}{\mbox{${\cal R(L)} \:$}}
\newcommand{\La}{\mbox{${\cal L}\:$}}
\newcommand{\A}{\mbox{${\cal A}\:$}}
\newcommand{\I}{\mbox{${\cal I}\:$}}
\newcommand{\R}{\mbox{${\cal R}\:$}}
\newcommand{\ASS}{\mbox{${\sf ASS}\:$}}
\newcommand{\gr}{\mbox{$\stackrel{r}{\longrightarrow}\:$}}
\newcommand{\cs}{\mbox{$\stackrel{c}{\longrightarrow}\:$}}
\newcommand{\then}{\item[$\Longrightarrow$]}

\title{Quantitative Constraint Logic Programming 
for Weighted Grammar Applications\thanks{I am greatly indebted to Steven
    Abney, Thilo G\"otz and Paul King for their valuable comments on
    this paper. Furthermore, I would like to thank Graham Katz, Frank
    Morawietz and two anonymous LACL referees for their
    helpful suggestions.}}

\author{Stefan Riezler}

\institute{Graduiertenkolleg ILS, Seminar f\"ur Sprachwissenschaft,\\
Universit\"at T\"ubingen, Wilhelmstr. 113, 72074 T\"ubingen, Germany.\\
{\small \tt riezler@sfs.nphil.uni-tuebingen.de}}

\begin{document}

\maketitle

\begin{abstract}
Constraint logic grammars provide a powerful formalism for expressing complex
logical descriptions of natural language phenomena in exact
terms. Describing some of these  phenomena may,
however, require some form of graded distinctions which are not
provided by such grammars. Recent approaches to weighted constraint
logic grammars attempt to address this issue by adding numerical
calculation schemata to the deduction scheme of the underlying CLP
framework.

Currently, these extralogical extensions are not related to the
model-theoretic counterpart of the operational semantics of CLP, i.e.,
they do not come with a formal semantics at all.

The aim of this paper is to present a clear formal semantics for
weighted constraint logic grammars, which abstracts away
from specific interpretations of weights, but nevertheless gives
insights into the parsing problem for such
weighted grammars.
Building on the formalization of constraint logic grammars in the CLP
scheme of \cite{HuS:88}, this formal semantics will be given by a
quantitative version of CLP. Such a quantitative CLP scheme can also be
valuable for CLP tasks independent of grammars.
\end{abstract}

\section{Introduction}

Constraint logic grammars (CLGs) provide a powerful formalism for complex
logical description and efficient processing of natural language
phenomena. Linguistic description and computational practice may, however, often
require some form of graded distinctions which are not
provided by such grammars.

One such issue is the task of ambiguity resolution. This problem can
be illustrated for formal grammars describing a nontrivial domain of
natural language as follows: For such grammars every input of
reasonable length may receive a
large number of different analyses, many of which are not in accord with
human perceptions. Clearly there is a need to distinguish more plausible
analyses of an input from less plausible or even totally spurious
ones.

This problem has successfully been addressed by the use of weighted
grammars for disambiguation in regular and context-free
grammars. Weighted grammars assign numerical values, or weights,
to the structure-building components of the grammars and calculate the
weight of an analysis from the weights of the structural features that
make it up. The correct analysis is chosen from among the
in-principle possible analyses  by assuming the analysis with the
greatest weight to be the correct analysis. This approach also allows
parsing to be speeded up by pruning low-weighted subanalyses.

The idea of weighted grammars recently has been transferred to highly
expressive weighted CLGs by
\cite{Erbach:93a,Erbach:95} and \cite{Eisele:94}.
The approaches of Erbach and Eisele are
based on the feature-based constraint formalism CUF (\cite{Doerre:91,Doerre:93}),
which can be seen as an instance of the constraint logic programming (CLP)
scheme of \cite{HuS:88}. These approaches extend the
underlying formalism by assigning weights to program clauses, but differ
with respect to an interpretation of weights in a preference-based
versus probabilistic framework. Erbach calculates a preference
value of analyses from the preference values of the clauses used in
the analyses, whereas Eisele assigns application probabilities to
clauses from which a probability distribution over analyses is
calculated.

There is an obvious problem with these approaches, however.
Even if the formal foundation of the underlying framework is clear
enough, there is no well-defined semantics for the weighted
extensions. This means that these extralogical extensions of the deduction
scheme of the underlying constraint logic program are not related to
the model-theoretic counterpart of this operational semantics, i.e.,
they do not come with a formal semantics at all. This is clearly an
undesirable state of affairs. Rather, in the same way as CLGs allow
for a clear model-theoretic characterization of linguistic objects
coupled with the operational parsing system, one would prefer to
base a quantitative deduction system on a clear quantitative
model-theory in a sound and complete way.

The aim of this paper is to present a clear formal semantics for
weighted CLGs, which abstracts away from specific interpretations of weights, but gives
insight into the parsing problem for weighted CLGs.
Building on the formalization of CLGs in the CLP scheme of
\cite{HuS:88}, this formal semantics will be given by a
quantitative version of CLP. Such a quantitative CLP scheme can also be
valuable for CLP tasks independent of grammars.

Previous work on related topics has been confined to quantitative
extensions of conventional logic programming.
A quantitative deduction scheme based on a fixpoint semantics for sets
of numerically annotated conventional definite clauses was first
presented by van Emden in \cite{Emden:86}. In this approach numerical
weights are associated with definite clauses as a whole. 
The semantics of such quantitative rule sets is based upon concepts of
fuzzy set algebra and crucially deals with the truth-functional ``propagation'' of
weights across definite clauses.
Van Emden's approach initialized research into a now extensively explored area of
quantitative logic programming. For example, annotated logic
programming as introduced by
\cite{Sub:87} extends the expressive power of quantitative rule sets
by allowing variables and evaluable function terms as
annotations. Such annotations can be attached to components of the
language formula and come with more complex mappings as a foundation
for a multivalued logical semantics. Such extended theories are
interpreted in frameworks of lattice-based logics for generalized
annotated logic programming (\cite{Kifer:92}), possibilistic
logic for possibilistic logic programming (\cite{Dubois:91}) or logics
of subjective probability for probabilistic logic programming
(\cite{Ng:92,Ng:93}) and probabilistic deductive databases
(\cite{Laks:94,LaksSad:94}).

Aiming at a formal foundation of weighted CLGs in
a framework of quantitative CLP, we can start from the ideas developed
in the simple and elegant framework of \cite{Emden:86}, but transfer
them to the general CLP scheme of \cite{HuS:88}.
This means that the form of weighted CLGs under
consideration allows us to restrict our attention to numerical
weights associated with CLP clauses as a
whole. Furthermore, the simple concepts of fuzzy set algebra can also provide
a basis for an intuitive formal semantics for quantitative
CLP. Such a formal semantics will be sufficiently general in
that it is itself not restricted by a specific interpretation of weights.
Further extensions should be straightforward, but have to be deferred to
future work. Our scheme will straightforwardly transfer the nice properties of
the CLP scheme of \cite{HuS:88} into a quantitative version of CLP.

\section{Constraint Logic Programming and Constraint Logic Grammars}

Before discussing the details of our quantitative extension of CLP,
some words on the underlying CLP scheme and grammars formulated by
these means are necessary. In the following we will rely on the
CLP scheme of \cite{HuS:88}, which generalizes conventional logic
programming (see \cite{Lloyd:87}) and also the CLP scheme of \cite{JuL:86}
to a scheme of definite clause specifications over arbitrary
constraint languages.
A very general characterization of the concept of constraint language
can be given as follows.

\begin{definition}[\La] A constraint language \La consists of
\begin{enumerate}
\item an \La-signature, specifying the non-logical elements of
  the alphabet of the language,
\item a decidable infinite set \Xsf{VAR} whose elements are called variables,
\item a decidable set \Xsf{CON} of \La-constraints which are
  pieces of syntax with unknown internal structure,
\item a computable function \Xsf{V} assigning to every constraint
  $\phi \in \Xsf{CON}$ a finite set $\Xsf{V}(\phi)$ of variables, the
  variables constrained by $\phi$,
\item a nonempty set of \La-interpretations \Xsf{INT},
where each \La-interpretation $\I \in \Xsf{INT}$ is defined
w.r.t.\ a nonempty set \Xcal{D}, the domain of \I, and a set {\sf ASS}
of variable assignments ${\sf VAR} \rightarrow
{\cal D}$,
\item a function \denI{\cdot} mapping
  every constraint $\phi \in \Xsf{CON}$ to a set \denI{\phi} of
  variable assignments, the solutions of $\phi$ in \I.
\item Furthermore, a constraint $\phi$ constrains only the variables
  in $\Xsf{V}(\phi)$, i.e., if $\alpha \in \denI{\phi}$ and $\beta$ is
  a variable assignment that agrees with $\alpha$ on $\Xsf{V}(\phi)$,
  then $\beta \in \denI{\phi}$.
\end{enumerate}
\end{definition}

To obtain constraint logic programs, a given constraint language
\Xcal{L} has to be extended to a constraint language \Xcal{R(L)}
providing for the necessary relational atoms and propositional
connectives.

\begin{definition}[\Rl] A constraint language \Rl extending a
  constraint language \La is defined as follows:
\begin{enumerate}
\item The signature of \Rl is an extension of the signature of \La with a decidable set \Xcal{R} of
relation symbols and an arity function $\Xsf{Ar}: \Xcal{R} 
\rightarrow \bbbn$.
\item The variables of \Rl are the variables of \La.
\item The set of \Xcal{R(L)}-constraints is the smallest set s.t.
\begin{itemize}
\item $\phi$ is an \Xcal{R(L)}-constraint if $\phi$ is an
\Xcal{L}-constraint,
\item $r(\vec{x})$ is an \Xcal{R(L)}-constraint, called an {\bf
atom},  if $r \in \Xcal{R}$ is a relation symbol with arity n and
$\vec{x}$ is an n-tuple of pairwise distinct variables,
\item $\emptyset$, F $\&$ G, $F \rightarrow G$ are 
\Xcal{R(L)}-constraints, if F and G are \Xcal{R(L)}-constraints,
\item $\phi$ $\&$ $B_1$ $\&$ $\ldots$ $\&$ $B_n \rightarrow A$  is 
an \Xcal{R(L)}-constraint, called a {\bf definite clause}, if $A, B_1,
\ldots, B_n$ are atoms and $\phi$  is an \Xcal{L}-constraint. We may
write a definite clause also as
$A \leftarrow \phi$ $\&$ $B_1$ $\&$ $\ldots$ $\&$ $B_n$.
\end{itemize}
\item The variables constrained by an \Rl-constraint are defined as follows:
  If $\phi$ is an \La-constraint, then $\Xsf{V}(\phi)$ is defined as
  in \La;
$\Xsf{V}(r(x_1, \ldots, x_n)):= \{ x_1, \ldots, x_n \}$; 
$\Xsf{V}(\emptyset):= \emptyset$;
$\Xsf{V}(F \;\&\; G):= \Xsf{V}(F) \cup \Xsf{V}(G)$;
$\Xsf{V}(F \rightarrow G) := \Xsf{V}(F) \cup \Xsf{V}(G)$.
\item For each \La-interpretation \I,
  an \Xcal{R(L)}-interpretation \Xcal{A} is an extension of an
\Xcal{L}-interpretation \Xcal{I} with
relations $r^{\cal A}$ on the domain $\cal D$  of \Xcal{A} with
appropriate arity for every $r \in \Xcal{R}$ and
the domain of \Xcal{A} is the domain of \Xcal{I}.
\item For each \Rl-interpretation \A, for each \La-interpretation \I, \denA{\cdot} is a
  function mapping every \Rl-constraint to a set of
  variable assignments s.t.
\begin{itemize}
\item $\denA{\phi} = \denI{\phi}$ if $\phi$ is an 
\Xcal{L}-constraint,
\item $\denA{r(\vec{x})} = \{ \alpha \in \Xsf{ASS}|$ 
$\alpha(\vec{x}) \in r^{\cal A} \}$,
\item $\denA{\emptyset} = \Xsf{ASS}$,
\item $\denA{F \;\&\; G} = \denA{F} \cap \denA{G}$,
\item $\denA{F \rightarrow G} = (\Xsf{ASS} \setminus \denA{F}) \cup \denA{G}$.
\end{itemize}
\end{enumerate}
\end{definition}

A constraint logic program then is defined as a definite clause
specification over a constraint language.

\begin{definition}[definite clause specification] A definite clause
  specification $\Xcal{P}$ over a constraint language \La is a set of
  definite clauses from a constraint language \Rl extending \La.
\end{definition}

Relying on terminology well-known for conventional logic programming,
H\"oh-feld and Smolka's generalization of the key result of
conventional logic programming can be stated as follows:\footnote{
Further conditions for this generalization to hold are
decidability of the satisfiability problem, closure under variable
renaming  and closure under intersection for the constraint languages
under consideration.}
First, for every definite clause specification \Xcal{P} in the
extension of an arbitrary constraint
language \La, every interpretation of \La can be extended to a minimal
model of \Xcal{P}. Second, the SLD-resolution method for
conventional logic programming can be generalized to a sound and
complete operational semantics for definite clause specifications not restricted to Horn theories.
In contrast to \cite{JuL:86}, in this scheme constraint languages are not required to
be sublanguages of first order predicate logic and do not have to be
interpreted in a single fixed domain. This makes this scheme
usable for a wider range of applications. Instead, a constraint is
satisfiable if there is at least one interpretation in which it has a
solution. Moreover, such interpretations do not have to be solution
compact. This was necessary in \cite{JuL:86} to provide a sound and
complete treatment of negation as failure, which is not addressed in
\cite{HuS:88}.

The term constraint logic grammars expresses the connection between
CLP and constraint-based grammars. Constraint-based grammars 
allow for a clear model-theoretic characterization of linguistic objects by stating grammars as sets of
axioms of suitable logical languages. However, such
approaches do not necessarily provide an operational interpretation of
their purely declarative specifications. This may lead to problems
with an operational treatment of declaratively well-defined problems
such as parsing. CLP provides one possible approach to an operational
treatment of various such declarative frameworks by an embedding of
arbitrary logical languages into constraint logic programs. CLGs thus
are grammars formulated by means of a suitable logical language which
can be used as a constraint language in the sense of
\cite{HuS:88}.\footnote{Clearly, a direct definition of an operational
  semantics for specific constraint-based grammars  is possible and
  may even better suit the particular frameworks. However, such
  approaches have to rely directly on the syntactic properties of the
  logical languages in question. Under the CLP approach, arbitrary
  constraint-based grammars can receive a unique operational semantics
  by an embedding into definite clause specifications. The main
  advantage of this approach is the possibility to put
  constraint-based grammar processing into the well-understood
  paradigm of logic programming. This allows the resulting programs to
  run on existing architectures and to use well-known optimization
  techniques worked out in this area.}

For example, for feature based grammars such as HPSG (\cite{PuS:94}), a quite direct
embedding of a logical language close to that of \cite{Smolka:88} into
the CLP scheme of \cite{HuS:88} is done in the formalism CUF
(\cite{Doerre:91,Doerre:93}). This approach directly offers the
operational properties of the CLP scheme by simply redefining grammars
as constraint logic programs, but is questionable in losing the connection to the
model-theoretic specifications of the underlying
feature-based grammars.
A different approach is given by \cite{Goetz:95} where a compilation
of a logical language close to that of
\cite{King:94} into constraint logic programs is defined. This translation
procedure preserves important model-theoretic properties by generating
a constraint logic program ${\cal P(G)}$ from a feature-based grammar
${\cal G}$ in an explicit way.

The parsing/generation problem for CLGs then is as follows.
Given a program ${\cal P}$ (encoding a grammar) and a definite goal
$G$ (encoding the string/logical form we want to parse/generate from),
we ask if we can infer an answer $\varphi$
of $G$ (which is a satisfiable \La-constraint encoding an analysis)
proving the implication $\varphi
\rightarrow G$ to be a logical consequence of ${\cal P}$.
\section{Quantitative Constraint Logic Programming}
\subsection{Syntax and Declarative Semantics of Quantitative Definite Clause
  Specifications}
Building upon the definitions in \cite{HuS:88}, we can define the
syntax of a quantitative definite clause specification \Pf very quickly.
A definite clause specification \Xcal{P} in \Rl can be extended to a
quantitative definite clause specification \Pf in \Rl simply by adding
numerical factors to program clauses.

The following definitions are
made with respect to implicit constraint languages \La and \Rl.

\begin{definition}[\Pf] A quantitative definite clause specification
\Pf  in \Rl is a finite set of quantitative
  formulae, called quantitative definite clauses, of the form:
$\phi$ $\&$ $B_1$ $\&$ $\ldots$ $\&$
$B_n$ $\mbox{}_f\!\rightarrow A$, 
where $A$, $B_1, \ldots, B_n$ are \Rl-atoms,
$\phi$ is an \La-constraint, $n \geq 0$, $f \in (0,1]$.
We may write a quantitative formula also as
$A \leftarrow_f \phi$ $\&$ $B_1$ $\&$ $\ldots$ $\&$
$B_n$.
\end{definition}
Such factors should be thought of as abstract weights which receive a concrete
interpretation in specific instantiations of \Pf by weighted CLGs.

In the following the notation \Rl will be used more generally to
denote relationally extended constraint languages which possibly
include quantitative formulae of the above form.

To obtain a formal semantics for \Pf, first we have to introduce an
appropriate quantitative measure into the set-theoretic specification
of \Rl-interpretations. One possibility to obtain quantitative
\Rl-interpretations is to base the set algebra of \Rl-interpretations
on the simple and well-defined concepts of fuzzy set algebra (see
\cite{Zadeh:65}).

Relying on H\"ohfeld and Smolka's specification of  base equivalent \Rl-interpretations,
i.e., \Rl-interpretations extending the same \La-interpretation, in terms
of the denotations of the relation symbols in these interpretations,
we can ``fuzzify'' such interpretations by regarding the denotations
of their relation symbols as fuzzy subsets of the set of tuples in the
common domain.

Given constraint languages \La and \Rl,  we interpret each n-ary relation symbol
$r \in \R$ as a fuzzy subset of ${\cal D}^n$,
for each \Rl-interpretation \A with domain \Xcal{D}.
That is, we identify the denotation of $r$ under \A with a total function:
$\mu(\_ \:  ; r^{\cal A}): {\cal D}^n \rightarrow [0,1]$, which can be
thought of as an abstract membership function.
Classical set membership is coded in this context by membership
functions taking only 0 and 1 as values.

Next, we have to give a model-theoretic characterization of
quantitative definite clauses. Clearly, any monotonous mapping could
be used for the model-theoretic specification of the interaction of
weights in quantitative definite clauses and accordingly for the
calculation of weights in the proof-theory of quantitative CLP. For
concreteness, we will instantiate such a mapping to the specific case
of Definition \ref{mod} resembling \cite{Emden:86}'s mode of rule
application. This will allow us to state the proof-theory of
quantitative CLP in terms of min/max trees which in turn enables
strategies such as alpha/beta pruning to be used for efficient
searching. However, this choice is not crucial for the substantial
claims of this paper and generalizations of this particular
combination mode to specific applications should be straightforward,
but are beyond the scope of this paper.

The following definition of model corresponds to the definition of
model in classical logic when considering only clauses with
$f = 1$ and mappings ${\cal D}^n \rightarrow \{ 0,1 \}$.

\begin{definition}[model] \label{mod}
An \Rl-interpretation \A extending some 
  \La-interpre-\\
tation \I  is a model of
  a quantitative definite clause specification \Pf 
iff for each $\alpha \in \ASS$, for each quantitative formula
$r(\vec{x}) \leftarrow_f \phi$ $\&$ $q_1({\vec{x}}_1)$
$\&$ $\ldots$ $\&$ $q_k({\vec{x}}_k)$ in \Pf: If
$\alpha \in \denI{\phi}$, then
$\mu(\alpha(\vec{x}); r^{\cal A}) \geq f \times min \{
\mu (\alpha({\vec{x}}_j); q_j^{\cal A}) | $ $1 \leq j \leq k \}$.
\end{definition}

Note that the notation of an \Rl-interpretation \A will be used
more generally to include interpretations of quantitative
formulae. \Rl-solutions of a quantitative formula are defined as
$\denA{r(\vec{x}) \leftarrow_f \phi \;\&\; q_1({\vec{x}}_1)
\;\&\; \ldots \;\&\; q_k({\vec{x}}_k)} = \{ \alpha \in \ASS |$
If  $\alpha \in \denI{\phi}$, then
$\mu(\alpha(\vec{x}); r^{\cal A}) \geq f \times min \{
\mu (\alpha({\vec{x}}_j); q_j^{\cal A}) | \; 1 \leq j \leq k \} \}$.

The concept of logical consequence is defined as usual.

\begin{definition}[logical consequence] \label{logcons}
A quantitative formula
$r(\vec{x}) \leftarrow_f \phi$
is a logical consequence of a
quantitative definite clause specification \Pf iff for each
\Rl-interpretation \A, \A is a model of
\Pf implies that \A is a model of
$\{ r(\vec{x}) \leftarrow_f \phi \}$.
\end{definition}
Furthermore, we have that $r(\vec{x}) \leftarrow_f \phi$ is a logical
consequence of \Pf implies that
$r(\vec{x}) \leftarrow_{f'} \phi$ is a logical consequence of
\Pf for every $f' \leq f$.

A {\bf goal} $G$ is defined to be a (possibly empty) conjunction
of \Rl-atoms and \La-constraints.
We can, without loss of generality, restrict
goals to be of the form $r(\vec{x})$ $\&$ $\phi$, i.e., a (possibly
empty) conjunction of a single relational atom $r(\vec{x})$ and an
\La-constraint $\phi$. This is possible as
for each goal $G = r_1({\vec{x}}_1)$ $\&$ $\ldots$ $\&$
$r_k({\vec{x}_k})$ $\&$ $\phi$
containing more than one relational atom, we
can complete the program with a new clause $C = r({\vec{x}_1}, \ldots,
{\vec{x}_k}) \leftarrow_1 r_1({\vec{x}}_1)$ $\&$ $\ldots$ $\&$
$r_k({\vec{x}_k})$ $\&$ $\phi$ with $G$ as antecedent and
a new predicate with all variables in $G$ as arguments as consequent.
Submitting the new predicate $r({\vec{x}_1}, \ldots,
{\vec{x}_k})$ as query yields the same results as would be
obtained when querying with the compound goal $G$.

Given some \Pf and some goal $G$,
a  {\bf \Pf-answer} $\varphi$ of $G$ is defined to be a satisfiable
\La-constraint $\varphi$ s.t.\
$\varphi$ $\mbox{}_f\!\rightarrow G$ is a logical consequence of
\Pf. A quantitative formula
$\varphi$ $\mbox{}_f\!\rightarrow r(\vec{x})$ $\&$ $\phi$ is defined to
be a logical consequence of \Pf iff every model of \Pf is a
model of $\{ \varphi$ $\mbox{}_f\!\rightarrow r(\vec{x})$ $\&$
$\phi \}$.
An \Rl-interpretation \A is defined to be a a model of 
$\{ \varphi$ $\mbox{}_f\!\rightarrow r(\vec{x})$ $\&$ $\phi \}$ iff 
$\denA{\varphi} \subseteq \denA{\phi}$ and \A is a
model of $\{ r(\vec{x}) \leftarrow_f \varphi \}$.

Aiming to generalize the key result in the declarative semantics of CLP---
the minimal model semantics of definite clause specifications
over arbitrary constraint languages---to our quantitative CLP scheme,
first we have to associate a complete lattice of interpretations with
quantitative definite clause specifications.

Adopting Zadeh's definitions for set operations, we can define a
partial ordering on the set of base equivalent
\Rl-interpretations.
This is done by defining set operations on these interpretations with
reference to set operations on the denotations of
relation symbols in these interpretations. 
We get for all base equivalent \Rl-interpretations ${\cal A}, \; {\cal
  A'}$:
\begin{itemize}
\item $\A \subseteq \A'$ iff 
for each n-ary relation symbol $r \in \R$,
for each $\alpha \in \ASS$,
for each $\vec{x} \in {\sf VAR}^n$:
$\mu(\alpha(\vec{x}); r^{\cal A}) \leq
\mu(\alpha(\vec{x}); r^{\cal A'})$,
\item $\A = \bigcup X$ iff 
for each n-ary relation symbol $r \in \R$,
for each $\alpha \in \ASS$,
for each $\vec{x} \in {\sf VAR}^n$:
$\mu(\alpha(\vec{x}); r^{\cal A}) = 
sup \{ \mu(\alpha(\vec{x}); r^{\cal A'}) | \; \A' \in X \}$,
\item $\A = \bigcap X$ iff 
for each n-ary relation symbol $r \in \R$,
for each $\alpha \in \ASS$,
for each $\vec{x} \in {\sf VAR}^n$:
$\mu(\alpha(\vec{x}); r^{\cal A}) = 
inf \{ \mu(\alpha(\vec{x}); r^{\cal A'}) | \; \A' \in X \}$,
\item $sup \; \emptyset = 0$, $inf \; \emptyset = 1$.
\end{itemize}
Clearly, the set of all base equivalent \Rl-interpretations is a
{\bf complete lattice} under the partial ordering of set inclusion.

Next we have to apply the syntactic notions of renaming and variant to
the quantitative case. 
 A {\bf renaming} is a bijection $\Xsf{VAR} \rightarrow \Xsf{VAR}$
 which is the identity except for finitely many exceptions and
\Xsf{VAR} is a decidable infinite set of variables.\\
A quantitative formula $\kappa'$ is a {\bf $\rho$-variant} of a
quantitative formula  $\kappa$ under a renaming $\rho$ iff
$\Xsf{V}(\kappa') =\rho(\Xsf{V}(\kappa))$, where \Xsf{V} is a computable
function assigning to every quantitative formula $\kappa$ the set
$\Xsf{V}(\kappa)$ of variables occurring in $\kappa$;
$\kappa' = \kappa\rho$, i.e., $\kappa'$ is the quantitative formula
obtained from $\kappa$ by simultaneously replacing each occurrence of a
variable $X$ in $\kappa$ by $\rho(X)$ for all variables in
$\Xsf{V}(\kappa)$;
and $\denA{\kappa} = \denA{\kappa'}\rho := \{ \alpha \circ
\rho |$ $\alpha \in \denA{\kappa'} \}$ for each interpretation \A.\\
A quantitative formula $\kappa'$ is a {\bf variant} of a quantitative
formula $\kappa$ if there exists a renaming $\rho$ s.t.\ $\kappa'$ is a
$\rho$-variant of $\kappa$.

Using these definitions, we
can state the central equations which link the declarative and procedural
semantics of \Pf.

\begin{definition} \label{equations}
Let \Pf be a quantitative definite clause specification in \Rl, \I
be an \La-interpretation. Then the countably infinite sequence $\left<
  {\cal A}_0, {\cal A}_1, {\cal A}_2, \ldots \right>$ of
\Rl-interpretations extending \I is a \Pf-chain iff for each n-ary
relation symbol $r \in \R$, for each $\alpha \in \ASS$, for each
$\vec{x} \in {\sf VAR}^n$:
\begin{description}
\item[]$\mu (\alpha(\vec{x}); r^{{\cal A}_0}) := 0$,
\item[]$\mu (\alpha(\vec{x}); r^{{\cal A}_{i+1}}) :=
max \{ f \times min \{
\mu (\alpha({\vec{x}}_j); q_j^{{\cal A}_i}) | $
$1 \leq j \leq n \}$ 
 $|$  there is a variant 
$r(\vec{x}) \leftarrow_f \phi$ $\&$ $q_1({\vec{x}}_1)$
$\&$ $\ldots$ $\&$ $q_n({\vec{x}}_n)$ of a clause in \Pf 
and $\alpha \in \denAi{\phi}{i} \}$.
\end{description}
\end{definition}

Before stating the central theorem concerning the declarative
semantics of quantitative definite clause specifications, we have to prove the
following useful lemma (cf. \cite{Emden:86}, Lemmata 2.10', 2.11'):

\begin{lemma}\label{supremum}
For each \Pf, for each \Pf-chain
$\left<{\cal A}_0, {\cal A}_1, {\cal A}_2, \ldots \right>$,
for each n-ary relation symbol $r \in \R$, for each  $\alpha \in
\ASS$, for each $\vec{x} \in {\sf VAR}^n$,
there exists some $n \in \bbbn$ s.t.\ $\mu(\alpha(\vec{x});
 r^{\bigcup_{i \geq 0} {\cal A}_i}) = \mu(\alpha(\vec{x}); r^{{\cal A}_n})$.
\end{lemma}

\begin{proof} We have to show that the supremum
$v = sup \{ \mu(\alpha(\vec{x}); r^{{\cal A}_i}) | \; i \geq 0 \} $ can be attained
for some $n \in \bbbn$.

\begin{description}
\item[$v=0$:] For $v = 0$, we have $n = 0$.

\vspace{1ex}

\item[$v > 0$:] For $v > 0$, we have to show for any real $\epsilon$,
  $0 < \epsilon <v$: $\{ \mu(\alpha(\vec{x}); r^{{\cal A}_i})|\; i \geq
  0 \; and \; \mu(\alpha(\vec{x}); r^{{\cal A}_i}) \geq \epsilon \}$ is
  finite.

Let $F$ be the finite set of real numbers of factors of clauses
  in \Pf, $m$ be the greatest element in $F$ s.t.\ $m < 1$ and let $q$ be
  the smallest integer s.t.\ $m^q < \epsilon$.\\
Then, since each real number
$\mu(\alpha(\vec{x}); r^{{\cal A}_i})$ is
a product of a sequence of elements of $F$, the number of different
products $\geq \epsilon$ is not greater than $|F|^q$ (in
combinatorics' talk, the permutation of $|F|$ different things taken $q$
at a time with repetitions) and thus finite.\\
Hence, the supremum is the maximum attained for some $n \in \bbbn$.
\qed
\end{description} 
\end{proof}

Now we can obtain minimal model properties for quantitative definite clause
specifications similar to those for the non-quantitative case of
\cite{HuS:88}. Theorem \ref{definite} states that we can construct a
minimal model \A of \Pf for each quantitative definite clause
specification \Pf in the extension of an arbitrary constraint language
\La and for each \La-interpretation. This means that---due to
the definiteness of \Pf---we can restrict our attention to a minimal model semantics of \Pf.

\begin{theorem}[definiteness] \label{definite}
For each quantitative definite clause specification \Pf in \Rl,
for each \La-interpretation \I,
for each \Pf-chain
$\left<{\cal A}_0, {\cal A}_1, {\cal A}_2, \ldots \right>$ of
\Rl-interpretations extending some \La-interpretation \I:

\begin{description}
\item[(i)] ${\cal A}_0 \subseteq {\cal A}_1 \subseteq \ldots$,
\item[(ii)] the union $\A := \bigcup_{i \geq 0} {\cal A}_i$ is a model
  of \Pf extending \I,
\item[(iii)] \A is the minimal model of \Pf extending \I.
\end{description}
\end{theorem}

\begin{proof} {\bf (i)} We have to show that 
${\cal A}_i \subseteq {\cal A}_{i+1}$.
We prove by induction on $i$ showing for each constraint language \La,
for each quantitative definite clause specification \Pf in \Rl, for
each \La-interpretation \I, for each \Pf-chain 
$\left<{\cal A}_0, {\cal A}_1, {\cal A}_2, \ldots \right>$ of
\Rl-interpretations extending some \La-interpretation \I,
for each n-ary relation symbol $r \in \R$, for each  $\alpha \in
\ASS$, for each $\vec{x} \in{\sf VAR}^n$, for each $i \in \bbbn$:
$\mu(\alpha(\vec{x}); r^{{\cal A}_i}) \leq \mu(\alpha(\vec{x});
r^{{\cal A}_{i+1}})$.

\begin{description}
\item[\rm Base:] $\mu(\alpha(\vec{x}); r^{{\cal A}_0}) = 0 \leq
  \mu(\alpha(\vec{x}); r^{{\cal A}_1})$.

\vspace{1ex}

\item[\rm Hypothesis:] Suppose $\mu(\alpha(\vec{x});r^{{\cal A}_{n-1}}) \leq
  \mu(\alpha(\vec{x});r^{{\cal A}_{n}})$.

\vspace{1ex}

\item[\rm Step:] $\mu(\alpha(\vec{x});r^{{\cal A}_{n}}) = v > 0$

\vspace{1ex}

\begin{description}
\then there exists a variant $r(\vec{x}) \leftarrow_f \phi$ $\&$
$q_1({\vec{x}}_1)$ $\&$ $\ldots$ $\&$ $q_k({\vec{x}}_k)$ of a clause
in \Pf s.t.\  $v = f \times min
\{ \mu(\alpha({\vec{x}_1});{q_1}^{{\cal A}_{n-1}}), \ldots,
\mu(\alpha({\vec{x}_k});{q_k}^{{\cal A}_{n-1}}) \}$ and $\alpha\in
\denAi{\phi}{n-1}$, by Definition \ref{equations}

\vspace{1ex}

\then $\mu(\alpha({\vec{x}_1});{q_1}^{{\cal A}_{n}}) \geq
\mu(\alpha({\vec{x}_1});{q_1}^{{\cal A}_{n-1}}), \\
\ldots,
\mu(\alpha({\vec{x}_k});{q_k}^{{\cal A}_{n}}) \geq
\mu(\alpha({\vec{x}_k});{q_k}^{{\cal A}_{n-1}})$ and
$\alpha\in \denAi{\phi}{n}$, by the hypothesis

\vspace{1ex}

\then $\mu(\alpha({\vec{x}});r^{{\cal A}_{n+1}}) \geq v$, by
definition of $\mu(\alpha({\vec{x}});r^{{\cal A}_{i+1}})$
\end{description}

\vspace{1ex}

\then $\mu(\alpha({\vec{x}});r^{{\cal A}_{n}}) \leq 
\mu(\alpha({\vec{x}});r^{{\cal A}_{n+1}})$.

\vspace{1ex}

\item[\rm For] $v=0$ follows immediately
$\mu(\alpha({\vec{x}});r^{{\cal A}_{n}}) \leq 
\mu(\alpha({\vec{x}});r^{{\cal A}_{n+1}})$.

\vspace{1ex}
\item[]Claim (i) follows by arithmetic induction.
\end{description}
{\bf (ii)} We have to show that $\A := \bigcup_{i \geq 0} {\cal A}_i$ is a
model of \Pf extending \I. We prove that for each clause
$r(\vec{x}) \leftarrow_f \phi$ $\&$
$q_1({\vec{x}}_1)$ $\&$ $\ldots$ $\&$ $q_k({\vec{x}}_k)$ in \Pf,
for each $\alpha \in \ASS$:
If $\alpha \in \denA{\phi}$, then
$\mu(\alpha(\vec{x}); r^{\cal A}) \geq f \times min \{
\mu(\alpha({\vec{x}_j}); {q_j}^{\cal A}) | \; 1 \leq j \leq k \}$.

\begin{description}
\item[] Note that since every ${\cal A}_i$ is an \Rl-interpretation
  extending \I, \A is an \Rl-interpretation extending \I.

\vspace{1ex}

\item[] Now let $r(\vec{x}) \leftarrow_f \phi$ $\&$
$q_1({\vec{x}}_1)$ $\&$ $\ldots$ $\&$ $q_k({\vec{x}}_k)$ be a clause
in \Pf s.t.\ for some $\alpha \in \ASS$:
$\alpha \in \denA{\phi}$ and $\mu(\alpha({\vec{x}_i});
{q_i}^{\cal A}) = min \{ \mu(\alpha({\vec{x}_j}); {q_j}^{\cal A}) | \;
1 \leq j \leq k \} = v$.

\vspace{1ex}

\item[]Then there exists some $n \in \bbbn$ s.t.\ $v =
\mu(\alpha({\vec{x}_i}); {q_i}^{{\cal A}_n}) =
min \{ \mu(\alpha({\vec{x}_j}); {q_j}^{{\cal A}_n}) | $\\
$ 1 \leq j \leq k \} $,
by Lemma \ref{supremum} and
since for all $j$ s.t. $1 \leq j \leq k: 
\mu(\alpha({\vec{x}_j}); {q_j}^{{\cal A}}) =
sup \{ \mu(\alpha({\vec{x}_j}); {q_j}^{{\cal A}_i}) | \; i \geq 0 \}$

\vspace{1ex}

\begin{description}
\then $\mu(\alpha({\vec{x}}); {r}^{{\cal A}_{n+1}}) \geq f \times v$,
by Definition \ref{equations}

\vspace{1ex}

\then $\mu(\alpha({\vec{x}}); {r}^{{\cal A}}) \geq 
\mu(\alpha({\vec{x}}); {r}^{{\cal A}_{n+1}})$, since 
$\mu(\alpha({\vec{x}}); {r}^{{\cal A}}) =$\\
$sup \{ \mu(\alpha({\vec{x}}); {r}^{{\cal A}_i}) | \; i \geq 0 \}$
\end{description}

\vspace{1ex}

\item{} $\Longrightarrow \mu(\alpha({\vec{x}}); {r}^{{\cal A}}) \geq f \times min
\{ \mu(\alpha({\vec{x}_j}); {q_j}^{\cal A}) | \;1 \leq j \leq k \}$.

\vspace{1ex}

\item[] This completes the proof for claim (ii).
\end{description}
{\bf (iii)} We have to show that \A is the minimal model of \Pf extending
\I. We prove for every base
equivalent model ${\cal B}$ of \Pf: ${\cal A}_i \subseteq {\cal B}$,
which gives $\A \subseteq {\cal B}$, by induction on $i$
showing for each constraint language \La,
for each quantitative definite clause specification \Pf in \Rl, for
each \La-interpretation \I, for each \Pf-chain 
$\left<{\cal A}_0, {\cal A}_1, {\cal A}_2, \ldots \right>$ of
\Rl-interpretations extending some \La-interpretation \I,
for each n-ary relation symbol $r \in \R$, for each  $\alpha \in
\ASS$, for each $\vec{x} \in{\sf VAR}^n$, for each $i \in \bbbn$:
$\mu(\alpha(\vec{x}); r^{{\cal A}_i}) \leq \mu(\alpha(\vec{x});r^{\cal
  B})$.

\begin{description}
\item[\rm Base:] $\mu(\alpha(\vec{x}); r^{{\cal A}_0}) = 0 \leq
\mu(\alpha(\vec{x}); r^{{\cal B}})$.

\vspace{1ex}

\item[\rm Hypothesis:] Suppose $\mu(\alpha(\vec{x}); r^{{\cal A}_{n-1}})
  \leq \mu(\alpha(\vec{x}); r^{{\cal B}})$.

\vspace{1ex}

\item[\rm Step:] $\mu(\alpha(\vec{x});r^{{\cal A}_{n}}) = v > 0$

\vspace{1ex}

\begin{description}
\then there exists a variant $r(\vec{x}) \leftarrow_f \phi$ $\&$
$q_1({\vec{x}}_1)$ $\&$ $\ldots$ $\&$ $q_k({\vec{x}}_k)$ of a clause
in \Pf s.t.\ $v = f \times min \{ \mu(\alpha({\vec{x}_1});{q_1}^{{\cal
    A}_{n-1}}), \ldots, \mu(\alpha({\vec{x}_k});{q_k}^{{\cal
    A}_{n-1}}) \}$ \\
and $\alpha \in
\denAi{\phi}{n-1}$, by Definition \ref{equations}

\vspace{1ex}

\then $\mu(\alpha({\vec{x}_1});{q_1}^{{\cal B}}) \geq
\mu(\alpha({\vec{x}_1});{q_1}^{{\cal A}_{n-1}}), \\
\ldots,
\mu(\alpha({\vec{x}_k});{q_k}^{{\cal B}}) \geq
\mu(\alpha({\vec{x}_k});{q_k}^{{\cal A}_{n-1}})$ and 
$\alpha \in \denX{\phi}{{\cal B}}$, by the hypothesis

\vspace{1ex}

\then $\mu(\alpha({\vec{x}});r^{{\cal B}}) \geq v$,
since ${\cal B}$ is a model of \Pf

\vspace{1ex}

\end{description}
\then $\mu(\alpha({\vec{x}});r^{{\cal A}_{n}}) \leq 
\mu(\alpha({\vec{x}});r^{{\cal B}})$.

\vspace{1ex}

\item[\rm For] $ v=0$ follows immediately
$\mu(\alpha({\vec{x}});r^{{\cal A}_{n}}) \leq 
\mu(\alpha({\vec{x}});r^{{\cal B}})$.

\vspace{1ex}

\item[] Claim (iii) follows by arithmetic induction.
\qed
\end{description}
\end{proof}

Proposition \ref{logicalcons}  allows us to link the declarative description of the
desired output from \Pf and a goal, i.e., a \Pf-answer, to
the minimal model semantics of \Pf. This is done by connecting the
concept of logical consequence with the concept of minimal model.

\begin{proposition} \label{logicalcons}
Let \Pf be a quantitative definite clause specification in
  \Rl, $\varphi$ be an \La-constraint and $G$ be a goal.
Then $\varphi$ $\mbox{}_v\!\rightarrow G$ is a
  logical consequence of \Pf iff every minimal model \A of \Pf
is a model of $\{ \varphi$ $\mbox{}_v\!\rightarrow G \}$.
\end{proposition}

\begin{proof}
\begin{description}
\item[\rm if:] For each minimal model \A of \Pf:
\Xcal{A} is a model of $\{ \varphi$ $\mbox{}_v\!\rightarrow G \}$

\begin{description}
\then for every base equivalent model \Xcal{B} of \Pf: \Xcal{B}
is a model of $\{ \varphi$ $\mbox{}_v\!\rightarrow G \}$,
since $\Xcal{A} \subseteq \Xcal{B}$ by Theorem \ref{definite}, (iii)

\vspace{1ex}

\then $\varphi$ $\mbox{}_v\!\rightarrow G$
is a logical consequence of \Pf.
\end{description}

\vspace{1ex}

\item[\rm only if:] $\varphi$ $\mbox{}_v\!\rightarrow G$
is a logical consequence of \Pf

\begin{description}
\then every model of \Pf is a model of $\{ \varphi$
$\mbox{}_v\!\rightarrow G \}$, by Definition \ref{logcons}

\vspace{1ex}

\then \Xcal{A} is a model of $\{ \varphi$ $\mbox{}_v\!\rightarrow G \}$.
\qed
\end{description}
\end{description}
\end{proof}

The following toy example will illustrate the basic concepts of the
declarative semantics of quantitative definite clause specifications.

\begin{example}\label{toyex}
Consider a simple program \Pf consisting of clauses {\sf 1}, {\sf 2}
and {\sf 3}. Let for the sake of the example be $\denI{ X=\phi
  \;\&\;  X=\psi} = \emptyset$ for each \La-interpretation \I.

\begin{quote}
{\tt 1} $ p(X) \leftarrow_{.7} X=\phi$.\\
{\tt 2} $ p(X) \leftarrow_{.5 } X=\phi$.\\
{\tt 3} $ p(X) \leftarrow_{.9 } X=\psi$.
\end{quote}

A \Pf-chain for predicate $p$ and an object $\alpha(X)$ allowed
by the \La-constraint $X=\phi$
is constructed as follows.

\begin{quote}
$\mu(\left< \alpha(X) \right> ; p^{{\cal A}_0}) = 0$,

$\mu(\left< \alpha(X) \right> ; p^{{\cal A}_1}) $
$= max \{ .7 \times min \; \emptyset, .5 \times min \; \emptyset \} $
$= .7 $,

$\mu(\left< \alpha(X) \right> ; p^{{\cal A}_2}) $
$= max \{ .7 \times min \; \emptyset, .5 \times min \; \emptyset \} $ 
$ =.7 $,

$\vdots$
\end{quote}

The membership value of this object in the denotation of $p$ under
the minimal model \A of \Pf is attained in step 1 and calculated as follows.

\begin{quote}
$\mu(\left< \alpha(X) \right> ; p^{\bigcup_{i \geq 0} {\cal A}_i})$ 
$= sup \{ 0, .7, .7, \ldots \} $
$= .7$ .
\end{quote}

Clearly, \A is a model of clauses {\tt 1} and {\tt 2}.
A similar calculation can be done for clause {\tt 3}.

\end{example}

\subsection{Operational Semantics of Quantitative Definite Clause
  Specifications}

The proof procedure for quantitative CLP is a search of a tree,
corresponding to the search of an SLD-and/or
tree in conventional logic programming and CLP. Such a tree is defined
with respect to the inference rules \gr and \cs of \cite{HuS:88} and a
specific calculation of node values. 
The structure of such a tree exactly reflects the construction of a
minimal model and thus may be defined as a min/max tree.
In the following we will assume implicit constraint languages \La and \Rl and
a given quantitative definite clause specification \Pf in \Rl. Furthermore, \Xsf{V} will
denote the finite set of variables in the query and the V-solutions of
a constraint $\phi$ in an interpretation \I are defined as
$\denIV{\phi} := \{ \alpha|_V |\; \alpha \in \denI{\phi} \}$ and
$\alpha|_V$ is the restriction of $\alpha$ to \Xsf{V}.

The first inference rule is given by a binary relation \gr, called
goal reduction, on the set of goals.
\begin{description}
\item[$A$ $\&$ $G$ $\stackrel{r}{\longrightarrow}$ $F$ 
$\&$ $G$]
if $A \leftarrow F$ is a variant of a clause in \Xcal{P}\\
s.t. $(\Xsf{V} \cup \Xsf{V}(G)) \cap \Xsf{V}(F) \subseteq \Xsf{V}(A)$.
\end{description}

A second rule takes care of constraint solving for the
\Xcal{L}-constraints appearing in subsequent goals. The rule takes the
conjunction of the \Xcal{L}-constraints from the reduced goal and the
applied clause and gives, via the black box of a suitable \Xcal{L}-
constraint solver, a satisfiable \Xcal{L}-constraint in solved form if the
conjunction of \Xcal{L}-constraints is satisfiable. 
The constraint solving rule can then be defined as a total function
$\stackrel{c}{\longrightarrow}$ on the set of goals.
\begin{description}
\item[$\phi$ $\&$ $\phi'$ $\&$ $G$ $\stackrel{c}{\longrightarrow}$ 
$\phi''$ $\&$ $G$]
if $ [ \! [  \phi \; \& \; \phi' ] \! ] ^{\cal I}
_{{\sf V} \cup {\sf V}(G)} = [ \! [   \phi'' ] \! ] ^{\cal I}
_{{\sf V} \cup {\sf V}(G)}$\\
for each \Xcal{L}-interpretation \Xcal{I} and for all \La-constraints
$\phi, \phi'$ and $\phi''$.
\end{description}

\begin{definition}[min/max tree] A min/max tree determined by a query $G_1$ and
a quantitative definite clause specification \Pf has to satisfy the
following conditions:
\begin{enumerate}
\item Each max-node is labeled by a goal.
The value of each nonterminal max-node is the maximum of the values of
its descendants.
\item Each min-node is labeled by a clause from \Pf and a goal.
The value of each nonterminal min-node is $f \times m$, where $f$ is
the factor of the clause and $m$ is the minimum of the values of its
descendants.
\item The descendants of every max-node are all min-nodes s.t.\ for
  every clause $C$ with \gr-resolvent $G'$ obtained by $C$ from goal
  $G$ in a max-node, there is a min-node descendant labeled by $C$
  and $G'$.
\item The descendants of every min-node are all max-nodes s.t.\ for
  every \Rl-atom $r(\vec{x})$ in goal
$G \: \& \: \phi \: \&\: \phi'$ in a min-node with \cs-resolvent 
$G \: \& \: \phi''$, there is a max-node descendant labeled by
$r(\vec{x}) \: \& \: \phi''$.
\item The root node is a max-node labeled by $G_1$.
\item A success node is a terminal max-node labeled by a satisfiable\\
  \La-constraint.
The value of a success node is 1.
\item A failure node is a terminal max-node which is not a success
  node.
The value of a failure node is 0.
\end{enumerate}
\end{definition}

\begin{definition}[proof tree]
A proof tree for goal $G_1$ from \Pf  is a subtree of a min/max
supertree determined by $G_1$ and \Pf and is defined as follows:
\begin{enumerate}
\item The root node of the proof tree is the root node of the
  supertree.
\item A max-node of the proof tree is a max-node of the supertree and
  takes {\em one} of the descendants of the supertree max-node as its
  descendant.
\item A min-node of the proof tree is a min-node of the supertree and
  takes {\em all} of the descendants of the supertree max-node as its
  descendants.
\item All terminal nodes in the proof tree are success nodes $\phi,\:
  \phi', \ldots$\\
 s.t.\ $\phi \;\&\; \phi'\;\&\;\ldots \cs \varphi$
and $\varphi$ is a satisfiable \La-constraint, called answer
constraint.
\item Values are assigned to proof tree nodes in the same way as to
  min/max tree nodes.
\end{enumerate}
\end{definition}

To prove soundness and completeness of this generalized SLD-resolution proof
procedure, some further concepts have to be introduced. 

First, we have to take care of {\bf renaming closure} of the
generalized constraint language \Rl. A constraint language is said to be closed
under renaming iff every constraint has a $\rho$-variant for every
renaming $\rho$. Clearly, \Rl is closed under renaming if the
underlying constraint language \La is closed under
renaming. Furthermore, for each \Rl closed under renaming, for each
\Rl-interpretation \A: \A is a model of an \Rl-constraint iff \A
is a model of each of its variants.

Next, we have to redefine a {\bf complexity measure} for goal
reduction for the quantitative case. This measure is crucial in
proving termination of goal reduction and works by keying steps of the
minimal model construction to steps of the goal reduction process.
\begin{itemize}
\item The complexity of a variable assignment $\alpha$ for an atom
$r(\vec{x})$ in the minimal model \A s.t.\
$\mu(\alpha(\vec{x}); r^{\cal A}) > 0$ 
is defined as $comp(\alpha, r(\vec{x}), \A)
:= min \{ i |$ $\mu(\alpha(\vec{x}); r^{\cal A}) =
\mu(\alpha(\vec{x}); r^{{\cal A}_i})\}$.
\item The complexity of $\alpha$ for goal $G = r_1(\vec{x}_1)$ $\&$ $\ldots$
$\&$ $r_k(\vec{x}_k)$ $\&$ $\phi$
in \A s.t.\ $\alpha \in \denA{\phi}$ and 
$\mu(\alpha(\vec{x}_i); {r_i}^{\cal A}) > 0$ for all $i: 1 \leq i \leq k$
is defined as $comp(\alpha,G,\A) := \{
comp(\alpha,r_i(\vec{x}_i),\A)|\; 1 \leq i \leq k \} $ where $ \{
\ldots \} $ is a multiset.
\item The V-complexity of $\alpha$ for goal $G=r_1(\vec{x}_1)$ $\&$ $\ldots$
$\&$ $r_k(\vec{x}_k)$ $\&$ $\phi$ in \A s.t.\
$\alpha \in \denAV{\phi}$ and 
$\mu(\alpha(\vec{x}_i); {r_i}^{\cal A}) > 0$ for all $i: 1 \leq i \leq k$
is defined as $comp_V(\alpha,G,\A) := min \{ comp(\beta,G,
\A) | $ $ \beta \in \denA{\phi}$,
$\mu(\beta(\vec{x}_i); {r_i}^{\cal A}) > 0$ for all $i: 1 \leq i \leq
k$ and $\alpha = \beta |_V \}$
where the minimum is taken with respect to a total ordering on
multisets s.t.\ $M \leq M'$ iff
$\forall x \in M \setminus M', \exists x' \in M'\setminus M$ s.t.\ 
$x < x'$.
\end{itemize}
Clearly, the constraint solving part of the deduction scheme does not
affect the denotation or complexity of subsequent goals.

The following proofs show that the quantitative proof procedure  is
sound and complete with respect to the above stated semantic concepts. Again,
there is a close similarity to the corresponding statements for the
non-quantitative case of \cite{HuS:88}.

\begin{theorem}[soundness] For each quantitative definite clause
  specification \Pf, for each goal $G$, for each \La-constraint
  $\varphi$:
If there is a proof tree for $G$ from \Pf with answer constraint
$\varphi$ and root value $v$, then
$\varphi$ $\mbox{}_v\!\rightarrow G$ is a logical consequence of \Pf.
\end{theorem}

\begin{proof} The result is proved by induction on the depth $d$
of the proof tree, where one unit of depth is from max-node to
max-node.

\begin{description}
\item[\rm Base:] We know that proof trees
  of depth $d=0$ have to take the form of a single max-node labeled by a satisfiable
  \La-constraint $\psi$ with root value 1.
Then $\psi$ $\mbox{}_1\!\rightarrow \psi$ is a logical
  consequence of \Pf.

\vspace{1ex}

\item[\rm Hypothesis:] Suppose the result holds for proof trees of depth $d <
  n$.

\vspace{1ex}

\item[\rm Step:] Let $G_0 = r(\vec{x}) \: \& \: \phi$
be a goal labeling a proof tree of depth $d = n$ with answer
constraint $\psi$ and root value $h$,\\
 let $G_0' =
q_1({\vec{x}}_1)$ $\&$ $\ldots$ $\&$ $q_k({\vec{x}}_k)$ $\&$
$\phi$ $\&$ $\phi'$ be a goal labeling the min-node
obtained from $G_0$ via \gr using the variant
$C' = r(\vec{x}) \leftarrow_f \phi'$ $\&$
$q_1({\vec{x}}_1)$ $\&$ $\ldots$ $\&$ $q_k({\vec{x}}_k)$ of a clause
$C$ in \Pf, \\
and let $G_1 = q_1({\vec{x}}_1)$ $\&$ $\phi'', \ldots,
G_k = q_k({\vec{x}}_k)$ $\&$ $\phi''$ be goals labeling max-nodes 
obtained from $G_0'$ via \cs.

\vspace{1ex}

Then each goal $G_1, \ldots, G_k$ labels a proof tree of depth $d <
n$ with respective answer constraint
$\psi_1, \ldots, \psi_k$ and root value
$g_1, \ldots, g_k$ s.t.\ $h = f \times min \{g_1, \ldots, g_k \}$ and
for each model \A of \Pf: $\denA{\psi} = \denA{\psi_1 \:\&\: \ldots \:\&\:
  \psi_k}$, by definition min/max tree

\vspace{1ex}

\begin{description}
\then $\psi_1$ $\mbox{}_{g_1}\!\rightarrow G_1, \ldots, 
\psi_k$ $\mbox{}_{g_k}\!\rightarrow G_k$ are logical consequences of
\Pf, by the hypothesis

\vspace{1ex}

\then for each model \A of \Pf, for each $\alpha \in \ASS$:
$\denA{\psi} \subseteq \denA{\phi''}$ and if
$\alpha \in \denA{\psi}$, then $\mu(\alpha(\vec{x}_1); {q_1}^{\cal A})
\geq g_1, \ldots, \mu(\alpha(\vec{x}_k); {q_k}^{\cal A})
\geq g_k$, by definition of logical consequence

\vspace{1ex}

\then  for each model \A of \Pf, for each $\alpha \in \ASS$:
$\denA{\psi} \subseteq \denA{\phi'}$ and if
$\alpha \in \denA{\psi}$, then $\mu(\alpha(\vec{x}); {r}^{\cal A})
\geq f \times min \{ \mu(\alpha(\vec{x}_1); {q_1}^{\cal A}), \ldots,
\mu(\alpha(\vec{x}_k); {q_k}^{\cal A}) \}$, since each model \A of \Pf is a model
of $C'$ iff \A is a model of $C$

\vspace{1ex}

\then for each model \A of \Pf, for each $\alpha \in \ASS$:
$\denA{\psi} \subseteq \denA{\phi}$ and if
$\alpha \in \denA{\psi}$, then $\mu(\alpha(\vec{x}); {r}^{\cal A})
\geq h$
\end{description}

\vspace{1ex}

\then $\psi$ $\mbox{}_{h}\!\rightarrow r(\vec{x})$ $\&$
$\phi$ is a logical consequence of \Pf.

\vspace{1ex}

\item[] The result follows by arithmetic induction.
\qed
\end{description}
\end{proof}

\begin{theorem}[completeness] Let \Pf be a quantitative definite clause
  specification in \Rl, \La be closed under renaming, \A be a
  minimal model of \Pf, $G$ be a goal of the form $r(\vec{x})$ $\&$
  $\phi$, $\alpha \in \denAV{\phi}$ and
$\mu(\beta(\vec{x}); r^{\cal A}) = v$ s.t.\ $v > 0$ and
$\alpha = \beta|_V$.
Then there exists a proof tree for $G$ from \Pf with answer constraint
$\varphi$ and root value $v$ and $\alpha \in \denAV{\varphi}$.
\end{theorem}

\begin{proof} The result is proved by induction on $c =
comp_V(\alpha, G, \A)$.

\begin{description}
\item[\rm Base:] We know that goals with complexity $c=\emptyset$ have
  to take the form of a satisfiable \La-constraint $\chi$.
Then there exists a proof tree for $\chi$ from \Pf consisting of a
single max-node labeled with $\chi$ and root value 1.

\vspace{1ex}

\item[\rm Hypothesis:] Suppose the result holds for goals with
  complexity $c < N$.

\vspace{1ex}

\item[\rm Step:] Let $G_0 = q(\vec{x})$ $\&$ $\psi$, $\alpha' \in
  \denAV{\psi}$, $\alpha'' \in \denA{\psi}$,
$\alpha' = \alpha''|_V$, $comp_V(\alpha',G_0,\A) $ $=$
$ comp(\alpha'',G_0,\A) = N$, $comp(\alpha'',q(\vec{x}),\A) := i$,
$\mu(\alpha''(\vec{x}); q^{\cal A}) = h$
and $h>0$.

\vspace{1ex}

\begin{description}
\item[] First we observe, that
$\mu(\alpha''(\vec{x}); q^{{\cal A}_i}) = h$,
since $comp(\alpha'',q(\vec{x}),\A) := i$
\vspace{1ex}

\begin{description}
\then there exists a variant
$q(\vec{x}) \leftarrow_f \psi'$ $\&$
$q_1({\vec{x}}_1)$ $\&$ $\ldots$ $\&$ $q_k({\vec{x}}_k)$ s.t.\ \\
$h= f \times min \{ \mu(\alpha({\vec{x}_1});
{q_1}^{{\cal A}_{i-1}}),$ $\ldots,$ 
$\mu(\alpha({\vec{x}_k});{q_k}^{{\cal A}_{i-1}}) \}$\\
and $\alpha'' \in \denAi{\psi'}{i-1}$
and $(\Xsf{V} \cup \Xsf{V}(\psi)) \cap
\Xsf{V}(\psi'$ $\&$ $q_1({\vec{x}}_1)$ $\&$ $\ldots$ $\&$
$q_k({\vec{x}}_k)) \subseteq \Xsf{V}(q(\vec{x}))$,
by definition \ref{equations} and renaming closure of \Rl, finite
\Xsf{V} and infinitely many variables in \Xsf{VAR}

\vspace{1ex}

\then $G_0 \stackrel{r,c}{\longrightarrow} G_0'$ s.t.\ $G_0' =
q_1({\vec{x}}_1)$ $\&$ $\ldots$ $\&$ $q_k({\vec{x}}_k)$ $\&$
$\psi''$\\
and $\denAV{\psi''} = \denAV{\psi \:\&\: \psi'}$, by definition
of the inference rules.
\end{description}

\vspace{1ex}

\item[] Next, $\alpha' \in \denAV{\psi''}$, since
$\alpha'' \in \denA{\psi}$, $\alpha'' \in \denAi{\psi'}{i-1}
\subseteq \denA{\psi'}$, \\
$\alpha'' \in \denA{\psi \:\&\: \psi'}$,
$\denAV{\psi \:\&\: \psi'} = \denAV{\psi''}$
and $\alpha' = \alpha''|_V$.

\vspace{1ex}

\item[] Finally, $comp_V(\alpha', G_0', \A) < N$, 
since $comp_V(\alpha', G_0', \A)$\\
$\leq$ $comp(\alpha'', G_0', \A)$
$<$ $\{ i\}$
$=$ $\{ comp(\alpha'', q(\vec{x}), \A) \}$
$=$ $comp(\alpha'',G_0, \A)$
$=$ $comp_V(\alpha',G_0,\A)$
$=$ $N$.

\vspace{1ex}

\item[] Now we can obtain goals $G_1 = q_1(\vec{x}_1)$ $\&$
  $\psi'',\ldots, G_k$ $
= q_k(\vec{x}_k)$ 
$\&$ $\psi''$ from $G_0'$
s.t.\ $\alpha' \in \denAV{\psi''}$, 
$\mu(\alpha''({\vec{x}_1}); {q_1}^{{\cal A}}) = g_1 > 0,
\ldots,$ 
$\mu(\alpha''({\vec{x}_k}); {q_k}^{{\cal A}}) $
$= g_k > 0$,\\
$\alpha' = \alpha''|_V$ and
$comp_V(\alpha',G_1,\A) < N, $\ldots 
$, comp_V(\alpha', G_k, \A) < N$.

\vspace{1ex}

\begin{description}
\then for each goal $G_1, \ldots, G_k$, there exists a proof tree from
\Pf with respective answer constraint $\chi_1, \ldots, \chi_k$ and
respective root value $g_1' = g_1, \ldots, g_k' = g_k$ and
$\alpha' \in \denAV{\chi_1 \:\&\: \ldots \:\&\:
\chi_k} = \denAV{\chi}$,
by the hypothesis
\end{description}
\end{description}

\vspace{1ex}

\then there exists a proof tree for $G_0$ from \Pf with answer
constraint $\chi$ and root value $h' = f \times min \{ g_1', \ldots,
g_k' \} = f \times min \{g_1, \ldots, g_k \} = h$ and $\alpha' \in
\denAV{\chi}$.

\vspace{1ex}

\item[] The result follows by arithmetic induction.
\qed
\end{description}
\end{proof}

Returning to our toy example, the proof procedure for quantitative
definite clause specifications can be illustrated as follows.

\begin{example} \label{toyexcont}
Starting from the simple program of Example \ref{toyex}, a min/max
tree for query $p(X)\;\&\; X=\phi$ and \Pf is constructed
as follows.

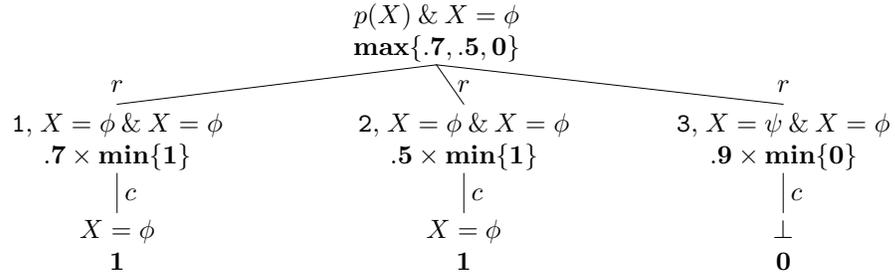
\begin{center}
\setlength{\GapWidth}{100pt}

\begin{bundle}
{$\begin{array}{c}
p(X) \:\&\:  X= \phi \\
\mathbf{ max \{ .7, .5, 0 \} }
\end{array}
$}

\chunk[$r$]{
\begin{bundle}
{$\begin{array}{c}
{\tt 1,} \;
 X=\phi \:\&\: X= \phi \\
\mathbf{ .7 \times min \{ 1 \} }
\end{array}
$}
\chunk[$\quad c$]
{$\begin{array}{c}
X= \phi \\
\mathbf{ 1}
\end{array}
$}
\end{bundle}}

\chunk[$r$]{
\begin{bundle}
{$\begin{array}{c}
{\tt 2,} \;
X= \phi \:\&\: X= \phi\\
\mathbf{ .5 \times min \{ 1\} }
\end{array}
$}
\chunk[$\quad c$] 
{$ \begin{array}{c}
X= \phi  \\
 \mathbf{ 1}
\end{array}
$}
\end{bundle}}

\chunk[$r$]{
\begin{bundle}
{$\begin{array}{c}
{\tt 3,} \;
X= \psi \:\&\: X= \phi\\
 \mathbf{  .9 \times min \{ 0 \} }
\end{array}
$}
\chunk[$\quad c$] 
{$ \begin{array}{c}
\bot \\
 \mathbf{ 0}
\end{array}
$}
\end{bundle}}

\end{bundle}

\end{center}

This tree contains two success branches and one failure branch (from
left to right). The proof trees obtained from this min/max tree are as
follows.

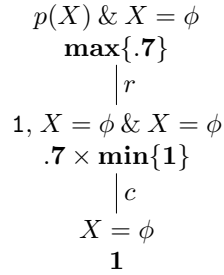
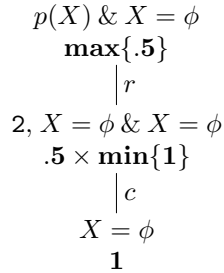
\begin{center}

\begin{bundle}
{$\begin{array}{c}
p(X) \:\&\:  X=\phi \\
\mathbf{ max \{ .7 \} }
\end{array}
$}

\chunk[$\quad r$]{
\begin{bundle}
{$\begin{array}{c}
{\tt 1,} \;
X= \phi \:\&\: X= \phi \\
\mathbf{ .7 \times min \{ 1 \}}
\end{array}
$}
\chunk[$\quad c$]
{$\begin{array}{c}
X= \phi \\
\mathbf{ 1}
\end{array}
$}
\end{bundle}}

\end{bundle}

\end{center}

\vspace{1ex}

\begin{center}

\begin{bundle}
{$\begin{array}{c}
p(X) \:\&\:  X=\phi \\
\mathbf{ max \{ .5 \} }
\end{array}
$}

\chunk[$\quad r$]{
\begin{bundle}
{$\begin{array}{c}
{\tt 2,} \;
X= \phi \:\&\: X= \phi \\
\mathbf{ .5 \times min \{ 1 \} }
\end{array}
$}
\chunk[$\quad c$]
{$\begin{array}{c}
X= \phi \\
\mathbf{ 1}
\end{array}
$}
\end{bundle}}

\end{bundle}

\end{center}

Clearly, $X=\phi \; \mbox{}_{.7}\!\rightarrow p(X) \;\&\;
X=\phi$ is a logical consequence of \Pf.
\end{example}

As proposed by \cite{Emden:86}, search strategies such as alpha-beta
pruning (see \cite{Nilsson:82}) can be used quite directly to define
an interpreter for quantitative rule sets. The same techniques can be
applied to a min/max proof procedure in quantitative CLP. 
In general, the amount of search
needed to find the best proof for a goal, i.e., the maximal valued
proof tree for a goal from a program, will be
reduced remarkably by controlling the search by the alpha-beta
algorithm.
 
\section{Quantitative CLP and Weighted CLGs}

To sum up, the quantitative CLP scheme presented above allows for a definition
of the parsing problem (and similarly of the
generation problem) for weighted CLGs in the following way:
Given a program $\Xcal{P}_F$ (coding some weighted CLG) and a query $G$
(coding some input string), we ask if we can infer a \Pf-answer
$\varphi$ of $G$ (coding an analysis)
at a value $\upsilon$ (coding the weight of the analysis) proving
$\varphi \;\mbox{}_{\upsilon}\!\rightarrow G$ to be a logical consequence
of \Pf.
The concept of weighted logical consequence thus can be seen as a
model-theoretic counterpart to the operational concept of weighted
inference.

We showed soundness and completeness results for a general proof
procedure for quantitative constraint logic programs with respect to a
simple declarative semantics based on concepts of fuzzy set
algebra. These terms in turn allow for a deeper characterization of the
concept of weighted logical consequence:
A \Pf-answer to a query $G=r(\vec{x}) \;\&\; \phi$
at value $\upsilon$ is a satisfiable \La-constraint $\varphi$ such
that for each model \A of \Pf holds:
If $\varphi$ is satisfiable, then $\phi$ is satisfiable and all
objects assigned to $\vec{x}$ by a solution of $\varphi$ are in the
denotation of $r(\vec{x})$ at a membership value of at least
$\upsilon$.

Considering concrete instantiations and applications of this formal
scheme, the remaining question is how to give the concept of
weight an intuitive interpretation. In the following we will briefly
discuss two possible interpretations of weighted CLGs each
of which is determined by the specific aims of a specific application.

One interpretation of weights is as a correlate to the degree of grammaticality
of an analysis. In \cite{Erbach:93a,Erbach:95}, Erbach attempts to
calculate the degree of grammaticality of an analysis from the
application probabilities of clauses used in the analysis and additional user-defined
weights.\footnote{Erbach sketches a calculation scheme which employs a
  restricted summation over clause weights instead of a minimization as
  is done in our quantitative CLP scheme. This calculation scheme
  could easily be captured by our quantitative CLP scheme by replacing
  $min$ by a restricted sum in the relevant definitions of the
  declarative semantics and accordingly of the procedural semantics of
  our scheme.}
Regardless of the motivation for this specific
determination of degrees of grammaticality, the choice to interpret weights  in correspondence
to degrees of grammaticality severely restricts the possible
applications of such weighted CLGs.

Considering for example the problem of ambiguity
resolution which is also addressed by Erbach, we think that the
concepts of preference value and degree
of grammaticality should be clearly differentiated.
As discussed in \cite{Abney:96}, the problem of ambiguity resolution
cannot be reduced to some few unrealistic examples. Instead, when describing a nontrivial part of
natural language, grammars of the usual sort will produce massive
artificial ambiguity where we can find
grammatical readings even for the most abstruse analyses.
Suppose for example a grammar which licenses, among many others,
analyses such as {\em  1) John
believes $[$Peter saw Mary$]_S$} and
{\em 2) John believes $[$$New$ $York_N$
$taxi_N$ $drivers_N$$]_{NP}$}. Such a grammar would also
license the analysis {\em 3) John believes $[Peter_N$ $saw_N$ $Mary_N
  ]_{NP}$} (provided a noun entry for the noun reading of {\em saw}),
which is clearly less preferred than {\em 1)}. Analysis {\em 3)} otherwise
is not less grammatical, as we can find an
acceptable reading (where the $NP$ refers to the $Mary$ associated
with some kind of $saw$ called a $Peter$ $ saw$). Degrading the weight of
the rule  $NP \rightarrow N$ $N$ $N$ (licensing multiple nominal
modifications) would on the other hand also degrade the weight of {\em
  2)}, which prevents a disambiguation by an interpretation of weights
in terms of degrees of grammaticality.

Considering the problem of graded grammaticality, it seems necessary to
employ richer models for a determination of degrees of grammaticality.
A first attempt to incorporate degrees of grammaticality investigated
by psycholinguistic experiments into CLGs is presented in
\cite{KellerDA:96,KellerKonv:96}.\footnote{Keller concentrates on
  experimental investigation of degrees of grammaticality and sketches
  a model of graded grammaticality based on ranked constraints. Such a model
  should easily be given a formal basis in terms of our quantitative
  CLP scheme.} Weighted CLGs interpreted in a
serious framework of graded grammaticality then
could provide a valuable framework for a clear
procedural and declarative treatment of graded grammaticality in CLGs.

Another interpretation of weighted CLGs is possible from the viewpoint
of probabilistic grammars. This approach has been shown to be
fruitful, e.g., for the problem of ambiguity resolution. The simple
but useful approximation adopted here is to assume the most plausible
analysis of a string to be the most probable analysis of that string.

An attempt to transfer the techniques of probabilistic context-free
grammars (see \cite{Booth:73}) to CLGs was presented
in \cite{Eisele:94}. In this approach the derivation process of CLGs
is defined as a stochastic process by the following stochastic model:
Each program clause gets assigned an application probability and the
probabilities of all clauses defining one
predicate have to sum to 1. The probability of a proof tree is
calculated as the product of the probabilities of the rules used in
it.\footnote{This calculation scheme also could easily be captured by
  our quantitative CLP scheme by replacing $min$ by a product
  accordingly in the relevant definitions of the declarative and
  procedural semantics of our scheme.} 
In order to make the probabilities of proof trees as defined by
the stochastic model constitute a proper probability distribution,
an additional normalization with respect to the overall probability of
proof trees has to be made.\footnote{\cite{Booth:73} discuss further
  conditions on consistency of probabilistic grammars which would have
  to be satisfied also by a probabilistic CLG model.}

What is interesting about probabilistic language models is their
ability to estimate the probabilistic parameters of the model in
accord to empirical probability distributions. 
Eisele attempts to estimate the probability of clauses proportional
to the expected frequency of clauses in derivations. Unfortunately,
this approach to parameter estimation is incorrect when
applied to the probabilistic CLG model of Eisele.
This means that the probability distribution over proof trees as defined
by a probabilistic CLG model estimated by the expected clause
frequency method is not in accord with the frequency of the proof
trees in the training corpus. Similarly, when dealing with unparsed
corpora, the EM-algorithm used for parameter estimation 
optimizes the wrong function when applied to this model.
The reason for this incorrectness is that the
set of trees generated from such a probabilistic CLG model is
constrained in a way which violates basic assumptions made in the applied
parameter estimation method. In other words, the probabilistic CLG
model defined by Eisele could be said to be incorrect, in the sense
that it makes an independency assumption for clause applications which
is violated by the languages generated from such probabilistic CLGs.

Since the proposed parameter estimation method is only provably
correct for the context-free case, the probabilistic language model of
Eisele faces a severe restriction. The only approach we know of to present a correct
parameter estimation algorithm for probabilistic grammars involving
context-dependencies is the model of stochastic attribute-value
grammars of \cite{AbneySAVG:96}, a discussion of which is beyond our
present scope.

\section{Conclusion}

We presented a simple and general scheme for quantitative
CLP.
Our quantitative extension straightforwardly transferred the nice
properties of the CLP scheme of \cite{HuS:88} into close analogs
holding for a quantitative version of CLP. With respect to
related approaches to quantitative extensions of conventional logic
programming, our extension raises ideas from this area to the general
framework of CLP.

We showed soundness and completeness results with respect to a
declarative semantics based on concepts of fuzzy set algebra. This
approach to a declarative semantics was motivated by the aim to give
a clear and simple formal semantics for weighted CLGs.

Clearly, more expressive quantitative extensions of CLP are possible
and will be addressed in future work. Regarding the interest in
computational linguistics problems such as ambiguity
resolution, however, a necessary prerequisite for a more sophisticated
semantics for probabilistically interpreted quantitative CLP is the
development of a probabilistic model for CLP which allows for correct
parameter estimation from empirical data.

\bibliographystyle{plain}

\end{document}